\documentclass[letterpaper]{article} 
\usepackage[submission]{aaai23}  
\usepackage{times}  
\usepackage{helvet}  
\usepackage{courier}  
\usepackage[hyphens]{url}  
\usepackage{graphicx} 
\usepackage{natbib}  
\usepackage{caption} 
\usepackage{algorithm}
\usepackage{algpseudocode}
\usepackage{algorithmicx}
\usepackage{newfloat}
\usepackage{listings}
\DeclareCaptionStyle{ruled}{labelfont=normalfont,labelsep=colon,strut=off} 
\floatstyle{ruled}
\newfloat{listing}{tb}{lst}{}
\floatname{listing}{Listing}
\usepackage{subfigure}

\usepackage{kotex}
\usepackage{xcolor}
\usepackage{multirow}
\title{Adaptive Attention Link-based Regularization for Vision Transformers}
\author {
    Heegon Jin,\textsuperscript{\rm 1}
    Jongwon Choi \textsuperscript{\rm 2,*}
}
\affiliations {
    \textsuperscript{\rm 1} Artificial Intelligence Graduate School, Chung-Ang University, Seoul, Korea\\
    \textsuperscript{\rm 2,*} Department of Advanced Imaging, Chung-Ang University, Seoul, Korea\\
    heegonjin@gmail.com\textsuperscript{\rm 1}\\
    choijw@cau.ac.kr\textsuperscript{\rm 2,*}
}

\usepackage{bibentry}

\begin{document}

\maketitle

\begin{abstract}
Although transformer networks are recently employed in various vision tasks with outperforming performance, extensive training data and a lengthy training time are required to train a model to disregard an inductive bias.
Using trainable links between the channel-wise spatial attention of a pre-trained Convolutional Neural Network (CNN) and the attention head of Vision Transformers (ViT), we present a regularization technique to improve the training efficiency of ViT.
The trainable links are referred to as the attention augmentation module, which is trained simultaneously with ViT, boosting the training of ViT and allowing it to avoid the overfitting issue caused by a lack of data.
From the trained attention augmentation module, we can extract the relevant relationship between each CNN activation map and each ViT attention head, and based on this, we also propose an advanced attention augmentation module.
Consequently, even with a small amount of data, the suggested method considerably improves the performance of ViT while achieving faster convergence during training.
\end{abstract}

\section{Introduction}

Convolutional Neural Networks~(CNN) have become standard for solving image-related tasks using deep neural networks since the advent of large publicly available datasets~\cite{DBLP:conf/nips/KrizhevskySH12,DBLP:conf/cvpr/HeZRS16}. 
Recently, the attention-based models studied in the area of natural language processing are becoming to take part in solving image-based tasks, which is called the Vision Transformer~(ViT)~\cite{DBLP:conf/iclr/DosovitskiyB0WZ21,DBLP:conf/iccv/TouvronCSSJ21}.
ViT is a transformer-based neural network fed by the patches of images with class-token for classification, replacing its input of the embedded words in natural language processing. 

Although ViT outperforms modern CNNs in terms of accuracy by ignoring the inductive bias of locality, a significant amount of data is required to train the model and achieve satisfactory performance without overfitting issues.
When we have an insufficient training data for ViT training, its performance becomes much worse than that of CNN due to the lack of inductive bias that can regularize the embedded parameters.
Furthermore, most researchers with limited computing resources are not affordable to train the ViT due to its lengthy training time and large parameters.

The overfitting and lengthy training issues must be solved to broaden the usability of ViT, so many recent studies have tried to solve the problems.
We can divide the studies by three categories: the advanced architecture-based method~\cite{DBLP:conf/iccv/TouvronCSSJ21,Yuan_2021_ICCV,DBLP:journals/corr/abs-2106-03714,DBLP:conf/iccv/LiuL00W0LG21}, the parameter compression-based method~\cite{DBLP:conf/iclr/ChoromanskiLDSG21,DBLP:journals/corr/abs-2006-04768}, and the knowledge distillation-based method~\cite{DBLP:conf/icml/TouvronCDMSJ21,DBLP:conf/iccv/CaronTMJMBJ21}.
The advanced architecture-based methods manipulate the architecture of ViT to achieve improved training efficiency and generalized prediction even with the small dataset.
However, the architectural manipulation of initial ViT~\cite{DBLP:conf/iclr/DosovitskiyB0WZ21} cannot be applied to different variants of ViTs, hence limiting their employment for new ViT-based models.
On the other hand, the parameter compression-based methods focus on a low-rank approximation of the self-attention mechanism in ViT, which results in reduced complexity but still suffers a computation-accuracy trade-off.

To overcome the limitations of the advanced architecture-based and parameter compression-based methods, the knowledge distillation-based methods utilize the predictions of well-trained models.
DeiT~\cite{DBLP:conf/icml/TouvronCDMSJ21} demonstrated meaningful development for a small dataset and reduced training time by employing label-based distillation.
Meanwhile, \cite{DBLP:journals/corr/abs-2006-00555} has argued that inductive bias can also be injected by knowledge distillation.
Although the methods can be used with minor model manipulations like a distillation token, it still has limitations in that the training datasets for both the student and teacher models must be equivalent.
This increases the training costs for training the teacher models before the knowledge distillation processes.

In this paper, we propose a novel regularization method of ViT models for reducing convergence time and avoiding overfitting on a small dataset. 
The proposed method utilizes an \textit{attention augmentation module} containing multiple trainable weights that estimate the affinity between the channel-wise activation map of CNN and the head-wise attention map of ViT.
Since the attention augmentation module is located outside of the ViT model, the ViT architecture can be perfectly preserved, allowing us to use the proposed algorithm in ViT variants based on the self-attention mechanism.
Furthermore, because the activation map of a CNN can be obtained regardless of its training dataset, the teacher CNN is not required to be trained on the same dataset as the student ViT.
The attention augmentation module and the teacher CNN model are only for the training of ViT, so our model preserves the number of weight parameters from the baseline models in the inference phase.

We validate our regularization method by using ImageNet, CIFAR-10, CUB-200, and Flowers-102 datasets with various scenarios, which shows the outperforming accuracy and the reduction of epochs required for its training convergence.
We also perform extensive ablation studies to show the effectiveness of our framework by varying data augmentation settings and distillation methods.
Furthermore, we analyze the trained weights of the attention augmentation module to investigate the factors for ViT to avoid the overfitting issue, and through the analysis, we present the dissimilarity of the deep layers' roles between CNN and ViT.

We can summarize our contributions as follows:
\begin{itemize}
\item We propose a novel regularization method to resolve the issues of overfitting and lengthy training time of ViT through the trainable attention links between the ViT attention maps and CNN activation maps.
\item The proposed scheme preserves the original architecture of ViT, which results in its general employment in various ViT models containing self-attention mechanism.
\item Through the proposed algorithm, the performance of ViT can be dramatically improved with the limited size of a dataset, and the training time is reduced without the loss of performance in various scenarios.
\item The relationship between ViT and CNN is analyzed in terms of attentional regions, which validates the analysis from the previous studies.
\end{itemize}

\section{Related Works}
\subsection{Transformers in Vision} 
Transformer models introduced by~\cite{DBLP:conf/nips/VaswaniSPUJGKP17} are neural networks that purely utilize the attention mechanism. 
While they have been used broadly in the field of natural language processing, Vision Transformer~(ViT)~\cite{DBLP:conf/iclr/DosovitskiyB0WZ21} adapted them in the domain of computer vision with minimal modification to its architecture.
ViT showed comparable performance to CNN in the condition of large pre-training.
For the advanced optimization of ViT, CaiT~\cite{DBLP:conf/iccv/TouvronCSSJ21} used layer normalization in ViT layers and changed the input location of class tokens to prevent saturation of performance in deep layers.
Swin Transformers~\cite{DBLP:conf/iccv/LiuL00W0LG21} adopted a hierarchical transformer that computes shifted windows to make it suitable for the vision domain.
PiT~\cite{DBLP:conf/iccv/HeoYHCCO21} introduced the concept of pooling in ViT from CNN, improving the generalization of ViT.
T2T-ViT~\cite{Yuan_2021_ICCV} enhanced sample efficiency by reshaping input tokens and changing the backbone of networks motivated by several CNN architectures.
Raghu et al.~\cite{DBLP:conf/nips/RaghuUKZD21} measured the similarity of representations between specific layers of CNN and ViT using centered kernel alignment.
With additional relative positional encoding, Cordonnier et al.~\cite{DBLP:conf/iclr/CordonnierLJ20} proved attention mechanisms in ViTs can perform as convolution layers in CNNs and showed their functional similarity.
From the investigation, ConViT~\cite{DBLP:conf/icml/dAscoliTLMBS21} was motivated to use relative positional encoding to give locality -- the inductive bias of CNNs -- to ViT.

The research was extended to \cite{DBLP:journals/corr/abs-2106-05795}, reparameterizing pre-trained convolutional layers as a format of ConViT. Refiner~\cite{DBLP:journals/corr/abs-2106-03714} tackled the over-smoothing problem between tokens in deep layers of ViT, and relieve it by projecting attention heads into the higher dimensions and applying convolution directly to attention maps to learn local relationship among the tokens. 
Those variants of ViT improved the optimization and data efficiency of the initial ViT model by modifying the architecture itself. 
However, our method does not touch any part of ViT modules, but only connects attention between ViT and CNN, transferring attention through links to give ViT a learning signal from the teacher.

\subsection{Knowledge Distillation} 
In knowledge distillation, a student model leverages a pre-trained teacher model's soft prediction divided by the same temperature values~\cite{DBLP:journals/corr/HintonVD15}.
The softened predictions can be regarded as label-smoothing, and by using them, the student model can achieve the data augmentation effect. 
Distillation between different types of neural architectures has also been proposed, DistilBERT~\cite{DBLP:journals/corr/abs-1903-12136} showed the effectiveness of a distilled knowledge from BERT~\cite{DBLP:conf/naacl/DevlinCLT19} into LSTM~\cite{DBLP:journals/neco/HochreiterS97}.
DeiT~\cite{DBLP:conf/icml/TouvronCDMSJ21} distilled knowledge from CNN to ViT, which seems similar to our work.
The knowledge distillation has been also employed in the transformer-based model of natural language processing, which results in performance improvement by using the teacher model~\cite{ref:nlp1, ref:nlp2}
However, in contrast to the previous study using the prediction for knowledge distillation, our framework transfers the knowledge based on the similarity of the latent feature maps.
As a result, we can extend the range of teacher models to cover the models in which the prediction vectors differ from the prediction of the task.

On the other hand, we can transfer latent representations of teacher models to those of student models.
FitNets~\cite{DBLP:journals/corr/RomeroBKCGB14} improved the stability of deep network training by guiding the latent layers to the teacher's well-trained latent representation.
Zagoruyko et al.~\cite{DBLP:conf/iclr/ZagoruykoK17} considered attention as projected activation maps of CNN into a spatial dimension, which could be regarded as spatial attention. They showed that spatial attention contains valuable information that is useful to improve the performance of the student network. 
Kim et al.~\cite{DBLP:conf/nips/KimPK18} used a paraphraser to extract and pass the teacher's knowledge to the student's translator to learn its representation.
Meanwhile, Heo et al.~\cite{DBLP:conf/aaai/HeoLY019a} demonstrated that the knowledge transfer based on the neurons' activation is a more classification-friendly approach than the direct transfer using output values.
Attention-based feature distillation~\cite{DBLP:conf/aaai/JiHP21} measured the similarities between teacher and student features through attention, which determines the importance of knowledge to transfer.

\begin{figure}[t]
\centering
  \includegraphics[width=1.\linewidth]{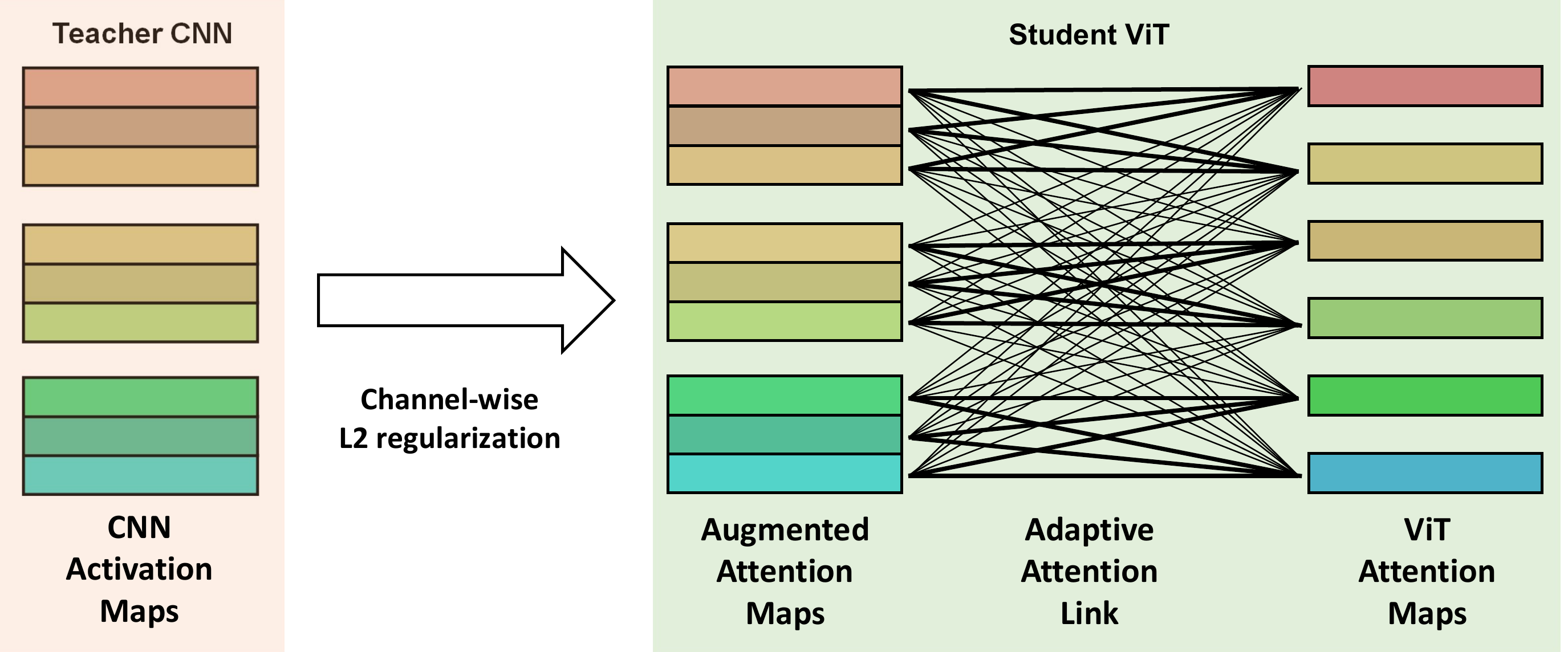}
\caption{{\textbf{Overall framework.} ViT attention maps are augmented through adaptive attention links, which build linear combinations of different original maps. Then, those augmented attention maps mimic CNN activation maps at one-on-one correspondence.}}
  \label{fig:framework}
\end{figure}

\section{Attention Link-based ViT Regularization}
In this section, we first explain the backgrounds of the self-attention mechanism and ViT.
Then, we explain the method to extract the attention maps from ViT, followed by the description of the architecture and the training method of the augmented attention module is described.
The overall framework is depicted in Fig.~\ref{fig:framework}.

\subsection{Background of ViT}
We first explain the self-attention mechanism and the original ViT model referred to by \cite{DBLP:conf/iclr/DosovitskiyB0WZ21}.
The self-attention mechanism mimics the human cognition system making the attention to the external stimulus, which is designed by a transformer-based model with the attention matrix estimated by pairs of key and query.

\subsubsection{Self-attention Mechanism}
We define the input sequence by $\mathbf{X}\in\mathcal{R}^{L\times D_{in}}$ where $L$ is the length of the sequence and $D_{in}$ means the dimension of one sequential element in the sequence.
Then, we can estimate the elements of the attention mechanism composed of key, query, and value vectors by linearly projecting the input sequence by the corresponding embedding weights $\mathbf{W}_k$, $\mathbf{W}_q$, and $\mathbf{W}_v$, respectively.
Thus, when we define the key, query, and value vectors by $\mathbf{Q}$, $\mathbf{K}$, and $\mathbf{V}$, respectively, we obtain the vectors as:
\begin{equation}
    \mathbf{K}=\mathbf{X}\mathbf{W}_k, \mathbf{Q}=\mathbf{X}\mathbf{W}_q, \mathbf{V}=\mathbf{X}\mathbf{W}_v,
\end{equation}
where $\mathbf{W} \in \mathcal{R}^{D_{in}} \times D_{head}$ from $\mathbf{W}\in\{\mathbf{W}_k, \mathbf{W}_q, \mathbf{W}_v\}$ and $D_{head}$ is the dimension of the head embedding.

Then, the self-attention of the head can be estimated by:
\begin{equation}
    f(\mathbf{X}) = s\Big(\mathbf{Q}\mathbf{K}^T / \sqrt{D_{head}}\Big)\mathbf{V}\in\mathcal{R}^{L\times D_{head}},
\end{equation}
where $s(\mathbf{Z})$ is a function to transfer each row vector of input matrix $\mathbf{Z}$ by softmax.
According to the derivation, self-attention can consider the semantic dependency among sequential inputs.

Many transformer-based models are based on the architecture stacked by the Multi-Head Self-Attention layers~(MHSA) containing multiple self-attention heads with independent embedding weights.
For the given input $\mathbf{X}$, we define the output of $m$-th self-attention head at $n$-th level depth by $f_{(m,n)}(\mathbf{X})$.
Then, we denote the corresponding key, query, and value vectors by $\mathbf{K}_{(m,n)}$, $\mathbf{Q}_{(m,n)}$, and $\mathbf{V}_{(m,n)}$, respectively.

\subsubsection{ViT Framework}
The original ViT model directly employed the conventional transformer-based model built for the natural language processing of the visual classification task.
We can summarize the inference process of the original ViT as following.
At first, we divide an input image by $P^2$ patches with the same size and sequentially order the patches after their vectorization.
Since the transformer network is invariant to the order of the sequential data, ViT concatenates positional embedding vectors to the input patches to represent the original position of the patch.

We define the sequential data obtained from one image by $\mathbf{X}_0\in\mathcal{R}^{P^2\times D_{im}}$, where $D_{im}$ is the size of the vector linearly projected from the vectorized image patch and the positional embedding vector.
Before we feed $\mathbf{X}_0$ into the transformer modules, the trainable class token sized by $\mathcal{R}^{D_{im}}$ is sequentially connected ahead of $\mathbf{X}'_0$, which results in $\mathbf{X}_0\in\mathcal{R}^{(P^2+1)\times D_{im}}$.

The transformer-based encoders of ViT are the modules containing the series of a layer normalization, a self-attention multi-head module, a fully connected layer, and a layer normalization, where every normalization layer has a residual connection.
We define the serial process by a function of $\mathbf{X}_{n+1}=g(\mathbf{X}_n)\in\mathcal{R}^{(P^2+1)\times D}$ where $D$ is the size of latent vectors.
When $N$ modules are stacked in the transformer-based encoder, the class-wise score is estimated by linearly projecting the final output of the class token as following: $p(cls|\mathbf{X}) = softmax(FC(\mathbf{X}_N))$.
For a detailed explanation of ViT, you can be referred to \cite{DBLP:conf/iclr/DosovitskiyB0WZ21}.

\begin{figure*}[t]
    \centering
    \subfigure[ImageNet]{\includegraphics[width=0.28\linewidth]{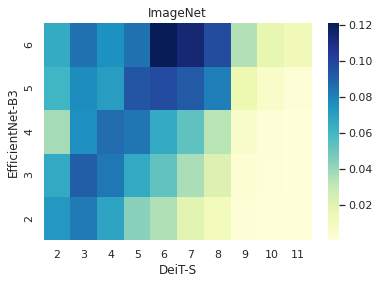}}\hfill
    \subfigure[ImageNet-5\%]{\includegraphics[width=0.28\linewidth]{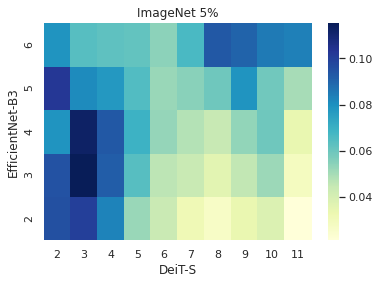}}\hfill
    \subfigure[Cifar-10]{\includegraphics[width=0.28\linewidth]{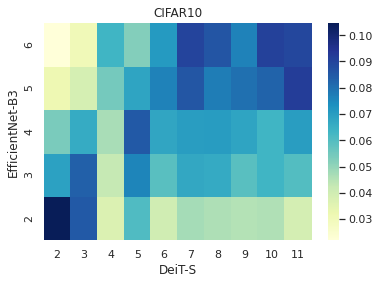}}
    \caption{\textbf{Relations between CNN Activation Maps and ViT attention maps.} The x-axis and y-axis indicates the depth of ViT layer and CNN block, respectively. We obtain the heat maps by averaging the magnitudes of attention links for respective level. The plots show the correlation along the layer depth, which is consistent with previous research~\cite{DBLP:conf/nips/RaghuUKZD21}. Furthermore, when a larger dataset (ImageNet) is trained, the link connectivity to high layers of student ViT decreases.}
    \label{fig:heatmaps}
\end{figure*}

\subsection{Attention Map Extraction}
We need to compare the ViT attention map and the CNN activation map for our regularization-based algorithm.
Instead of the relative positional embedding~\cite{DBLP:conf/icml/dAscoliTLMBS21} or the attention bias~\cite{DBLP:journals/corr/abs-2106-03714}, we preserve the original architecture of ViT to generalize the usability of our framework to cover the ViT variants.

To obtain the attention map from the original ViT, we utilize the attention value between the class token and the image patch.
The class token takes a key role to determine the final prediction, so we can assume that the attention to the class token may represent the importance of image patches for the classification result.
Thus, when feeding the class token into the transformer module as its query vector, we obtain the attention value by estimating the dot product between the embedding vectors of the class token and the corresponding image patch.
For $m$-th head in $n$-th multi-head self-attention layer, we can estimate the attention value as:
\begin{equation}
    \mathbf{A}_{(m,n)} = Rec\Big(s\left(\mathbf{Q}_{(m,n)}[0] \mathbf{K}_{(m,n)}[1:]^T\right)\Big)\in\mathcal{R}^{P\times P}
\end{equation}
where $Rec$ is a function to reconstruct the rectangular matrix of $\mathcal{P\times P}$ from its input vector of $\mathcal{P^2}$ according to the order of the sequential patches, and $\mathbf{Q}_{(m,n)}[0]$ and $\mathbf{K}_{(m,n)}[1:]$ represent the first query vector of the class token and the key vectors of the image patches, respectively.

In the case of the CNN activation map, we extract the activation maps after the normalization of every convolution block.
Instead of integrating the channel-wise activation maps, we consider the separated activation maps independently to improve the degree of freedom of our attention augmentation module.
In contrast to the constant resolution of ViT attention maps, the resolution of the CNN activation maps decreases with deep layers by pooling layers and strides of convolution layers.
Thus, to preserve the resolution of every activation map, we resize all the CNN activation maps to have the same size with the ViT attention maps by using bi-cubic interpolation.
We define the $c$-th resized CNN activation map by $\mathbf{B}_c$, where $c\in\{1,...,C\}$ and $C$ is the number of entire CNN activation maps.

\subsection{Attention Augmentation Module}

\subsubsection{Module Architecture}
Even though both the CNN activation and ViT attention maps represent the key parts of the target object for the prediction, their distribution such as a center point and a variance would be different from each other due to the dissimilarities of their operations.
For example, while the ViT attention map is distributed between 0 and 1 because of the softmax estimation, the values in the CNN activation map are normalized by a batch normalization, which can contain negative values.
Furthermore, in general, the number of CNN activation maps is much larger than the number of ViT attention maps due to the large channel-wise depth of CNNs.
Thus, it is impossible to directly compare each of the CNN activation maps with the ViT attention maps.

The attention augmentation module is designed to solve the problems of different distributions and a varying number of maps.
We design the attention augmentation module to contain multiple attention links which are the trainable weight parameters to scale the ViT attention maps.
By estimating the weighted summation of ViT attention maps with the attention links, we can obtain the augmented attention maps where the number is equivalent to the number of CNN activation maps.
Thus, we can estimate the augmented attention maps as following:
\begin{equation}\label{eq:qugattmap}
    \mathbf{A}^+_c = \sum_{m,n=1}^{M,N} w_c^{(m,n)} \mathbf{A}_{(m,n)} + b_c
\end{equation}
where $w_{c}^{(m,n)}$ and $b_c$ are the attention link and a trainable bias for $c$-th augmented attention map ($c\in\{1,...,C\}$), respectively.
$M$ is the number of self-attention heads in one level depth and $N$ presents the maximum level depth.
The size of $w_c$ is determined by the channel depth size of the CNN model and the number of self-attention heads in ViT model, which is irrelevant to the spatial size of the attention map.
Note that the weight of attention link $w_c^{(m,n)}$ is used to analyze the strength of connectivity for each CNN/ViT layer in link selection.

We implement the augmented attention module by a $1\times 1$ convolution layer generating $C$ augmented attention maps from a tensor of $\mathcal{R}^{P^2\times M N}$ where the ViT attention maps are stacked.
Because we use the augmented attention maps only for the training loss, the attention augmentation module has no role in the inference, which can be removed after the training of ViT.

\subsubsection{Module Training}
By using the augmented attention module, we can obtain the same number of augmented attention maps $\mathbf{A}^+_c$ with the CNN attention maps $\mathbf{B}_c$.
To ignore the remaining scale gap between the two maps, we first apply the $l2$ normalization, and then the mean squared error is estimated to build the attention-based regularization loss as:
\begin{equation}\label{eq:regatt}
    \mathcal{L}_{att} = \big\| {\mathbf{A}^+_c}/{\|\mathbf{A}^+_c\|_2} - {\mathbf{B}_c}/{\|\mathbf{B}_c\|_2}\big\|_2.
\end{equation}

Then, we integrate the attention-based regularization loss $\mathcal{L}_{att}$ with the cross-entropy loss $\mathcal{L}_{CE}$ of original ViT as:
\begin{equation}\label{eq:ceatt}
    \mathcal{L} = \mathcal{L}_{CE} + \lambda\mathcal{L}_{att},
\end{equation}
The adaptive attention link which is $w_c$ in Eq.~\ref{eq:qugattmap} is trained in the end-to-end scheme with Eq.~\ref{eq:ceatt}. 
Since $A_c^+$ is differentiable by $w_c$, $\mathcal{L}_{att}$ of Eq.~\ref{eq:regatt} directly updates $w_c$ to reduce the L2 distance between the CNN activation map and the augmented attention map. 
Meanwhile, $\mathcal{L}_{CE}$ also affects implicitly the update of $w_c$ due to its influence on $A$ in Eq.~\ref{eq:qugattmap}.

The term $\lambda$ is a scaling factor to control the effect of our regularization.
Since the regularization loss can work as an obstacle to ignoring the inductive bias, referred by \cite{DBLP:conf/icml/dAscoliTLMBS21}, we suppress the value of $\lambda$ at the specified epochs to increase the effectiveness of the cross-entropy loss. We exponentially decay the value of $\lambda$ by multiplying a decay rate at every epoch. 

\section{Link Selection for Advanced Regularization}
In this section, we build the advanced architecture of the attention augmentation module based on the analysis of the fully-trained  attention links.
After showing the resultant attention links, we explain the advanced link designed by considering the relations between CNN activation maps and ViT attention maps.

\begin{table}[t]
    \centering
    \resizebox{1.0\linewidth}{!}{
        \begin{tabular}{c|c|c|c|c|c|c|c|c}
            \hline
             \multicolumn{2}{c|}{CNN Block Level} & 1 & 2 & 3 & 4 & 5 & 6 & 7\\\hline
             \multirow{2}{*}{ViT Layer Level} & $\alpha$-link & 1-3 & 1-5 & 3-5 & 4-6 & 4-6 & 6-11 & 7-12 \\
             & $\beta$-link & 1-2 & 1-5 & 3-5 & 4-6 & 4-7 & 6-10 & 7-10\\
             \hline
        \end{tabular}
    }
    \caption{\textbf{Selective link configuration for student ViT (DeiT-S) and teacher CNN (EfficientNet-B3).} For example, the activation maps from the first CNN block are transferred to the ViT attention heads of the first to third layers. We obtain the selected links from the full link by using the structured link pruning procedure explained in Algorithm~\ref{algorithm} in supplementary material.}
    \label{tab:selLink}
\end{table}

\subsection{Analysis of Resultant Attention Links}
We visualize heat maps to show the scale distribution of the attention links after their training.
To compare the relationship between the ViT attention and the CNN activation maps, we only consider the magnitude of the weight parameters in the attention links.
As shown in Fig.~\ref{fig:heatmaps}, we obtain multiple heat maps by using three datasets including \textit{ImageNet}~\cite{ImageNet}, a 5\% subset of \textit{ImageNet}, and \textit{CIFAR-10}~\cite{Krizhevsky09learningmultiple}.

As analyzed in many previous studies~\cite{DBLP:conf/iclr/CordonnierLJ20, DBLP:conf/nips/RaghuUKZD21}, the ViT attention maps are highly related to the CNN activation maps selected at similar layer depths of the respective architecture.
The results validate that the multi-head self-attention of ViT can learn the hierarchical information by the stacked architecture, which is similar to the training mechanism of CNN.
Thus, we would be able to effectively regularize the ViT attention maps when we can handle their layer-level similarities with the CNN activation maps.

Furthermore, the heat maps from the attention links reveal an intriguing feature in which the attention heads at high layers of student ViT are no longer needed to be regularized when a large dataset is provided.
In comparison to small datasets, we can see suppressed magnitudes of attention links at high-level heads when a large dataset is provided.
To confirm the interesting point, we show the attention weight changes according to the training epochs in Fig.~\ref{Figure:attention link} in supplementary materials.
At the end of the training, we observed that high-level heads are disconnected from augmented attention maps, which means they cannot be effectively regularized by CNN activation maps.
Thus, the representation can be seen outside of CNNs' inductive bias, demonstrating that semantic information that overwhelms the inductive bias is difficult to train without a large dataset.

We can summarize the two hypotheses from the analysis as following:
\begin{itemize}
    \item The ViT attention and the CNN activation maps have correlation along with their layer depth.
    \item The attention heads at high layers of student ViT can present the semantic information that cannot be represented by the CNN layers, but training the semantic information requires a large dataset.
\end{itemize}

\subsection{Selective Attention Link}
Based on the analysis, we additionally propose the selective attention link to improve the training efficiency of the attention augmentation module.
Instead of the fully-connected link in the original attention augmentation module, only a part of the attention links are utilized to obtain the augmented attention maps.


Accordingly, we build two types of selective attention link, which are denoted by $\alpha$-link and $\beta$-link. 
Those selective links are obtained from pruning the full link by their connectivity, which are trained on subsampled ImageNet and full ImageNet, respectively.
$\alpha$-link connects the ViT attention maps to only the CNN activation maps at similar layer level.
$\beta$-link is similar to the $\alpha$-link but the links from CNN activation maps to the high-level heads are entirely disconnected.
The connections are given in Table~\ref{tab:selLink} and Algorithm~\ref{algorithm} of the supplementary material describes the detailed procedures to obtain the two selective attention links.

\section{Experiments}\label{sec:experiment}
\begin{table*}[t]
\centering
\resizebox{0.8\linewidth}{!}{
    \begin{tabular}{c||cccc|cccc}
    \hline
    \multirow{2}{*}{\begin{tabular}[c]{@{}c@{}}\\ Train Size\end{tabular}} & \multicolumn{4}{c|}{Top-1}                                                                                           & \multicolumn{4}{c}{Top-5}                                                                                           \\ \cline{2-9} 
                                                                           & \multicolumn{1}{c}{DeiT-S}   & \multicolumn{1}{c}{ConViT-S}          & \multicolumn{1}{c|}{AAL~(Ours)}       & \multicolumn{1}{c|}{Gap}         & \multicolumn{1}{c}{Deit-S}   & \multicolumn{1}{c}{ConViT-S}          & \multicolumn{1}{c|}{AAL(Ours)}       & \multicolumn{1}{c}{Gap}         \\ \hline\hline
    5\%                                                                    & \multicolumn{1}{c}{34.8\%} & \multicolumn{1}{c}{47.8\%}          & \multicolumn{1}{c|}{\textbf{56.1\%}} & 61\%/17\% & \multicolumn{1}{c}{57.8\%} & \multicolumn{1}{c}{70.7\%} & \multicolumn{1}{c|}{\textbf{80.0\%}}     & 38\%/13\%      \\ 
    10\%                                                                   & \multicolumn{1}{c}{48.0\%} & \multicolumn{1}{c}{59.6\%}          & \multicolumn{1}{c|}{\textbf{66.5\%}} & 39\%/12\% & \multicolumn{1}{c}{71.5\%} & \multicolumn{1}{c}{80.3\%}          & \multicolumn{1}{c|}{\textbf{87.4\%}} & 22\%/9\% \\ 
    30\%                                                                   & \multicolumn{1}{c}{66.1\%} & \multicolumn{1}{c}{73.7\%}          & \multicolumn{1}{c|}{\textbf{76.1\%}} & 15\%/4\% & \multicolumn{1}{c}{86.0\%} & \multicolumn{1}{c}{90.7\%}          & \multicolumn{1}{c|}{\textbf{93.0\%}} & 8\%/3\% \\ 
    50\%                                                                   & \multicolumn{1}{c}{74.6\%} & \multicolumn{1}{c}{78.2\%}          & \multicolumn{1}{c|}{\textbf{78.9\%}} & 6\%/1\%  & \multicolumn{1}{c}{91.8\%} & \multicolumn{1}{c}{93.8\%}          & \multicolumn{1}{c|}{\textbf{94.5\%}} & 3\%/1\% \\ 
    100\%                                                                  & \multicolumn{1}{c}{79.9\%} & \multicolumn{1}{c}{\textbf{81.4\%}} & \multicolumn{1}{c|}{81.0\%}          & 1\%/0\%  & \multicolumn{1}{c}{95.0\%} & \multicolumn{1}{c}{\textbf{95.8\%}} & \multicolumn{1}{c|}{95.5\%}          & 1\%/0\%  \\ \hline
    \end{tabular}
    }
\caption{\textbf{ImageNet test accuracy with different sampling ratios.} The \textit{Gap} columns represent our relative performance improvement over DeiT and ConViT, respectively. When the sub-sampling ratio is small, the performance gap enlarges, which indicates the performance of AAL to relieve over-fitting problem of ViT in the low-data regime. Meanwhile, with the large sub-sampling ratio, our model preserves the performance even while its model size is 22.3\% less than that of ConViT.}
\label{table:imagenet}
\end{table*}


In experiments, we show that transferring attention from pre-trained CNN models to ViTs can inject CNN's inductive bias~(i.e locality) naturally in standard self-attention layers, without the necessity of additional parameters extending the self-attention mechanism.
We call our proposed model as the Adaptive Attention Link~(AAL), and we examine how efficiently AAL helps ViT for improving its performance, especially showing a large gap in a small data regime.

\subsection{Experimental Settings}

\subsubsection{Implementation Details}
The computing resource used in our experiments is Nvidia A100. 
If not mentioned otherwise, the student ViT model used for experiments is DeiT-S (distilled version) and used EfficientNet-B3~\cite{DBLP:conf/icml/TanL19} as the teacher CNN model.
While our framework can utilize the teacher CNN models pre-trained by the datasets different from the ones for the student ViT models, we only consider the ImageNet pre-trained models as our teacher model. Thus, we can reduce the preparation time to re-train the new teacher model for the specific datasets.
For small sub-sampled ImageNet (5\% and 10\%), we apply batch size 512 which is the setting used for fine-tuning in original ViT~\cite{DBLP:conf/iclr/DosovitskiyB0WZ21}.
We set $\lambda$ to 2000 and the decay rate for $\lambda$ is set to 0.99 for the first 200 epochs and 0.98 for the last 100 epochs.
For a fair comparison, we preserve the values of the remaining hyperparameters and the training strategies from DeiT~\cite{DBLP:conf/icml/TouvronCDMSJ21}, our baseline.

\subsubsection{Comparisons and Dataset}
For comparison, we consider two previous studies, which include DeiT and ConViT.
The comparisons along the selected baselines are given in Table \ref{tab:comparison} of supplementary materials.
DeiT utilizes the knowledge distillation method to improve the ViT-based models, and ConViT shows state-of-the-art performance when the training data is given sufficiently even without using the knowledge distillation methods.
To show the generality of our algorithm, we utilize four classification datasets: ImageNet, CIFAR-10, Caltech-UCSD Birds-200-2011 (CUB-200)~\cite{WahCUB_200_2011}, and Oxford 102 Flowers (Flower-102)~\cite{Nilsback08}.
In the case of ImageNet, we extract the subsets randomly sampled with the various ratios (5\%, 10\%, 30\%, 50\%), maintaining class balance, to show the validity of the proposed algorithm when insufficient data is given for the training.

\subsubsection{Quantitative Results}
We first perform the comparisons with the various subsets of ImageNet.
As shown in Table~\ref{table:imagenet}, the proposed algorithm outperforms the state-of-the-art methods when the subsets of ImageNet  are used to train the model.
The performance of our framework is similar to that of ConViT when the entire dataset is considered for training. 
However, the performance gap between our framework and ConViT becomes enlarged with insufficient training data.
Furthermore, we should notice that our teacher model EfficientNet-B3 needs only 12.2M parameters, which is much smaller than 86.6M parameters of RegNetY-16GF~\cite{DBLP:conf/cvpr/RadosavovicKGHD20} used in DeiT~\cite{DBLP:conf/icml/TouvronCDMSJ21}.
Thus, we can validate that our proposed framework can overwhelm DeiT-B even by using the lighter teacher model.
In addition, while ConViT-S needs 5M more parameters than ours or DeiT-S, our method outperforms both of DeiT-S and ConViT-S, which validates the efficiency of our framework. 
We represent the quantitative results for CIFAR-10, CUB-200, and Flower-102 datasets in Tables~\ref{table:cifar} and \ref{tab:CUBflowers} of supplementary material.

\subsubsection{Training and Model Efficiency}

\begin{figure}[t]
\centering
        \includegraphics[width=0.9\linewidth,keepaspectratio]{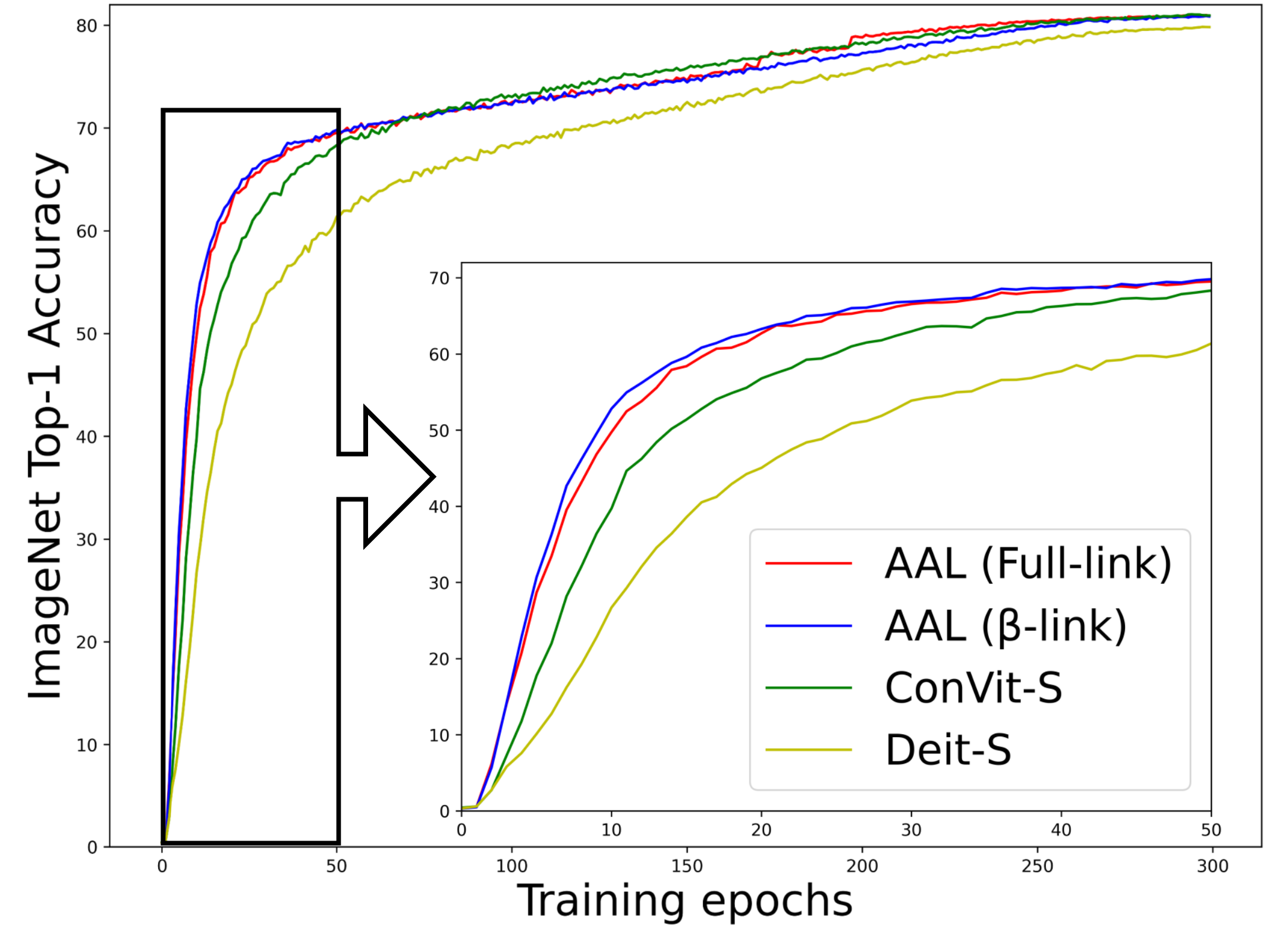}
\caption{\textbf{Comparison of learning curves.}}
    \label{Figure:learning curve}
\end{figure}


In Fig.~\ref{Figure:learning curve}, we compare the learning curve of our approach with those of DeiT and ConViT. 
Our method shows a larger performance gap than other models at the beginning stage of learning.
Also, the $\beta$-link which showed relevance between ViT and CNN on ImageNet stably boosts training with having prior connecting information compared to the full-link setting.
We can see that our algorithm achieves 70\% top-1 accuracy on ImageNet at about 50 epochs while the DeiT needs 120 epochs to reach the same accuracy. This result validates the rapid convergence of our approach, which comes from the correlation of the attention links in our analysis. 

Meanwhile, our additional trainable module, which is the attention augmentation module, includes only a single 1x1 Conv layer which augments the attention maps of the student ViT.
In our default settings, the number of the parameter is 0.068M, which is quite small compared to DeiT-S of 22M parameters.
Furthermore, the module can be removed after the training, so we can avoid the increase of running time and computational complexity in the inference phase.

\begin{table}[t]
\centering
\resizebox{1.0\linewidth}{!}{
    \begin{tabular}{c||cccc|cccc}
    \hline
    \multirow{2}{*}{\begin{tabular}[c]{@{}c@{}}\\Train  Size\end{tabular}} & \multicolumn{3}{c|}{Top-1} & \multicolumn{3}{c}{Top-5}  \\ \cline{2-7} 
                                                                           & \multicolumn{1}{c}{Full-link}   & \multicolumn{1}{c|}{Selective-link}& \multicolumn{1}{c|}{Gap}         & \multicolumn{1}{c}{Full-link}   & \multicolumn{1}{c|}{Selective-link} & \multicolumn{1}{c}{Gap}    \\ \hline\hline
    5\%                                                                    & \multicolumn{1}{c}{48.9\%} & \multicolumn{1}{c|}{\textbf{51.7\%}}          & \multicolumn{1}{c|}{{5.7\%}} & \multicolumn{1}{c}{73.6\%} & \multicolumn{1}{c|}{{\textbf{75.9\%}}} & \multicolumn{1}{c}{3.1\%}  \\ 
    10\%                                                                   & \multicolumn{1}{c}{63.0\%} & \multicolumn{1}{c|}{\textbf{64.7\%}}         & \multicolumn{1}{c|}{{2.6\%}} & \multicolumn{1}{c}{84.6\%} & \multicolumn{1}{c|}{\textbf{85.8\%}}         & \multicolumn{1}{c}{{1.4\%}}  \\ 
    30\%                                                                   & \multicolumn{1}{c}{75.2\%} & \multicolumn{1}{c|}{\textbf{76.1\%}}         & \multicolumn{1}{c|}{{1.2 \%}} & \multicolumn{1}{c}{92.4\%} & \multicolumn{1}{c|}{\textbf{93.0\%}}          & \multicolumn{1}{c}{{0.6\%}}  \\ 
    50\%                                                                   & \multicolumn{1}{c}{78.5\%} & \multicolumn{1}{c|}{\textbf{78.9\%}}          & \multicolumn{1}{c|}{{0.5\%}} & \multicolumn{1}{c}{94.3\%} & \multicolumn{1}{c|}{\textbf{95.0\%}}         & \multicolumn{1}{c}{{0.7\%}}\\ 
    100\%                                                                  & \multicolumn{1}{c}{\textbf{81.0\%}} & \multicolumn{1}{c|}{{80.9\%}} & \multicolumn{1}{c|}{-0.1\%}      &    \multicolumn{1}{c}{95.5\%} & \multicolumn{1}{c|}{\textbf{95.5\%}} & \multicolumn{1}{c}{0.0\%}   \\ \hline
    \end{tabular}
    }
\caption{\textbf{Comparison of full-link and selective link.} The \textit{Gap} columns represent the relative performance improvement of \textit{Selective-link}. For ablation, we maintain batch size 1024 at every experiment. For 100\% train size of ImageNet, we use $\beta$-link as the selective link, while $\alpha$-link is used for the remaining experiments. The selective link shows further improvement in test accuracy when the labeled data is scarce while not harming the performance on the large data.}
\label{table:link}
\end{table}

\begin{table}[t]
\centering
\resizebox{1.0\linewidth}{!}{
\begin{tabular}{c|c||cc}
\hline
Teacher Model & Student Model & \begin{tabular}[c]{@{}c@{}}Teacher Model\\ Top-1\end{tabular} & \begin{tabular}[c]{@{}c@{}}Student Model\\ Top-1\end{tabular} \\\hline\hline
ResNet34        & DeiT-S w/ distill & 73.3\% & \textbf{79.4}\% \\
EfficientNet-B3 & DeiT-B w/ distill & 81.1\% & \textbf{82.8}\% \\ \hline
\end{tabular}
}
\caption{\textbf{Ablation studies for teacher and student models.}}
\label{Table:Teacher/Student}
\end{table}



\subsection{Analysis}

In addition to the following analysis, we present the additional experiments to show the validity of our framework in the supplementary material.
The additional experiments include the performance of weakly supervised object localization, the qualitative results for attention maps, the learning curve, and epoch-wise qualitative changes of attention links.

\subsubsection{Effectiveness of Selective Links}
To show that our selective attention link-based transfer efficiently matches ViT attention maps with CNN activation maps, we compared two different settings on the attention augmentation module.
\textit{Full-link} fully connects each ViT attention map to produce augmented attention maps that match CNN activation maps as one-to-one channel-wise correspondence. 
In the case of the full ImageNet dataset, $\beta$-link was used for the selective link, while we utilized $\alpha$-link for the other small datasets. 

As shown in Table~\ref{table:link}, while the attention transfer with a fully connected attention link shows superior performance to the accuracy of DeiT and ConViT in a low data regime, the selective attention links show further improvement from its results in Table~\ref{table:link}.
The result also validates the analyzed correlation between the CNN activation map and the ViT attention map.
The weights of the full link are initialized by random values, so the training needs the initial computations to align the noisy weights letting the augmented attention maps be similar to the CNN activation map. In contrast, when we utilize the selective attention link, the noisy links can be ignored at the initial training phase, which results in reduced computation and stable training.

\subsubsection{Robustness to Variety of Models}
We add results with the variants composed of different teacher and student models to show the generality of our method in various environments.
As shown in Table~\ref{Table:Teacher/Student}, the proposed framework successfully improves the performance of its teacher model even with the different teacher and student models. Interestingly, with a light teacher model, we can achieve the large performance gap between the teacher and student models.


\begin{table}[t]
\centering
\resizebox{1.0\linewidth}{!}{
\begin{tabular}{ccc|clc}
\hline
                                             & \multicolumn{2}{c|}{Strong Data Aug.} & \multicolumn{3}{c}{Weak Data Aug.}         \\ \hline
\multicolumn{1}{c|}{Methods}                 & Top-1             & Top-5             & \multicolumn{2}{c}{Top-1}  & Top-5         \\ \hline\hline
\multicolumn{1}{c|}{Cross Entropy (CE)} & 91.3\%            & 99.6\%            & \multicolumn{2}{c}{84.2\%} & 98.7\%        \\
\hline
\multicolumn{1}{c|}{CE + AAL}                & \textbf{97.4}\%            & \textbf{99.9}\%            & \multicolumn{2}{c}{92.5\%} & 99.7\%        \\ 
\multicolumn{1}{c|}{CE + Soft Distillation}  & 91.0\%            & 99.6\%            & \multicolumn{2}{c}{84.0\%} & 98.9\%        \\
\multicolumn{1}{c|}{CE + Hard Distillation}  & 92.0\%            & 99.8\%            & \multicolumn{2}{c}{85.1\%} & 99.0\%        \\
\multicolumn{1}{c|}{CE + AAL + Hard Dist.}   & 96.5\%            & 99.9\%            & \multicolumn{2}{c}{\textbf{94.1\%}} & \textbf{99.7\%}        \\ \hline
\end{tabular}}

\caption{\textbf{Ablation studies for data augmentation and distillation methods on CIFAR-10.} To compare AAL with label-based distillation, we train our teacher model with CIFAR-10. We observe that AAL shows better performance than the label-based distillation methods, and we discover its synergy with label-based distillation upon the weak data augmentation.}
\label{Table: Ablation studies}
\end{table}


\subsubsection{Ablation Studies}

For additional verification of our knowledge transfer method, we perform experiments by using CIFAR-10 dataset with varying scenarios. 
To check the performance difference of knowledge distillation effect from each method, we reduce the effect of data augmentation from the setting used in DeiT by employing only simple augmentation methods such as random crops and horizontal flips.
This allows us to confirm the data efficiency in a low data regime.
In addition, we compared our method to other knowledge distillation methods introduced by DeiT, which uses a teacher model pre-trained on the CIFAR-10 dataset.
As shown in Table~\ref{Table: Ablation studies}, our method outperforms both the class prediction-based distillation methods using soft and hard labels.
From these results, we could discover that directly transferring attention is more effective for regularization than giving the teacher model's output predictions.

In addition, we can discover that the student ViT model cannot reach its best performance with the teacher model pre-trained by CIFAR-10.
This happens due to the low quality of the teacher model's intermediate representation, which would recall the advantage of AAL that can utilize the teacher models pre-trained by any dataset.
When AAL uses the teacher model pre-trained by ImageNet for the CIFAR-10 experiments, we acquire a large performance improvement, which is presented in Table~\ref{table:cifar} of supplementary.


\subsubsection{Various Baselines}
In Table~\ref{Table: PiT}, we show that applying our method is not only limited to standard ViT.
In the experiments, we employ our method to Pooling based ViT (PiT-S)~\cite{DBLP:conf/iccv/HeoYHCCO21}, and we observed the sample efficiency of the model increased by a large margin using our method.

\begin{table}[t]
    \begin{center}
    \resizebox{0.5\linewidth}{!}{%
        \centering
        \begin{tabular}{c|cc}
        \hline
        & Pit-S & Pit-S + AAL \\ \hline\hline
        Top-1 & 12.2\% & 44.0\% \\
        Top-5 & 25.2\% & 67.3\% \\ \hline
        \end{tabular}
        }
    \caption{\textbf{Comparison with PiT in ImageNet 5\%}}
    \label{Table: PiT}
    \end{center}
    \vspace{-2mm}

\end{table}


\section{Conclusion}
In this paper, we have introduced a novel method of transferring knowledge from CNN to ViT. By accessing attention of CNNs and adaptively adopting them, student ViT was able to earn high quality of learning signal with CNN's inductive bias. By applying our method, we could train ViT in less training epochs without overfitting even with the small dataset or limited labeled data. Also, we revealed  relations between intermediate representations from those different types of neural networks, which varied due to the training dataset. Furthermore, by analyzing those relationship with trained attention links, we take advantage of more efficient connection between networks.
\label{sec:conclusion}

\bibliography{aaai23}

\clearpage
{
\onecolumn
\centering
\Large
\textbf{Adaptive Attention Link-based Regularization for Vision Transformers} \\
\vspace{0.5em}Supplementary Material \\
\vspace{1.0em}
}
\appendix

\setcounter{table}{0}
\setcounter{figure}{0}
\renewcommand{\thetable}{\Alph{table}}
\renewcommand{\thefigure}{\Alph{figure}}

\begin{table}[ht!]
    \begin{center}
        \centering
        \begin{tabular}{c|ccc}
        \hline
        Models & DeiT-B & ConViT-S & AAL \\ \hline\hline
        Top-1  & 97.5\%   & 95.4\% & \textbf{97.5\%}   \\ \hline
        \end{tabular}
        \caption{CIFAR-10 Top-1 test accuracy}
        \label{table:cifar}
    \end{center}
\end{table}
Table~\ref{table:cifar} shows the experimental results with CIFAR-10 dataset.
The results verify that the proposed algorithm can increase the robustness to the insufficient size of training data since the model is randomly initialized.
In addition, in the DeiT paper, 7200 training epochs were needed to achieve 97.5\% top-1 test accuracy when training from scratch using the DeiT-B model which has more attention heads than DeiT-S.
On the other hand, our method needed only 300 training epochs to reach the same test accuracy as the DeiT-S model, which validates the training efficiency of our method.

\section{Fine-Grained Image Classification}
\begin{table}[ht!]
\centering
\begin{tabular}{c|cc|cc|cc|cc|}
\cline{2-9}
                                         & \multicolumn{2}{c|}{CUB-200 (AAL)} & \multicolumn{2}{c|}{CUB-200 (DeiT)} & \multicolumn{2}{c|}{Flowers-102 (AAL)} & \multicolumn{2}{c|}{Flowers-102 (DeiT)} \\ \hline
\multicolumn{1}{|c|}{Methods}            & \multicolumn{1}{c|}{Top-1}    & Top-5   & \multicolumn{1}{c|}{Top-1}        & Top-5        & \multicolumn{1}{c|}{Top-1}       & Top-5    & \multicolumn{1}{c|}{Top-1}          & Top-5          \\ \hline
\multicolumn{1}{|c|}{Scratch} & \multicolumn{1}{c|}{51.05\%}  & 80.03\% & \multicolumn{1}{c|}{26.49\%}      & 53.67\%      & \multicolumn{1}{c|}{94.13\%}     & 99.14\%    & \multicolumn{1}{c|}{87.16\%}        & 96.33\%        \\ \hline
\multicolumn{1}{|c|}{Transfer}  & \multicolumn{1}{c|}{84.76\%}  & 96.08\% & \multicolumn{1}{c|}{83.87\%}      & 95.87\%      & \multicolumn{1}{c|}{-}            &     -     & \multicolumn{1}{c|}{-}               &    -            \\ \hline
\end{tabular}
\caption{Classification accuracy on CUB-200 and Flowers-102.}
\label{tab:CUBflowers}
\end{table}

Since our method mainly considers the low-data regime, we regard fine-grained classification as another appropriate task to test our data efficiency.
We test classification accuracy on Caltech-UCSD Birds-200-2011 and Oxford Flowers-102, both of which have a small number of training samples per class.
In table~\ref{tab:CUBflowers}, `Scratch' row results are obtained by randomly initialized model, and `Transfer' row results by learning from the ImageNet pretrained model. 
We trained 300 epochs for both scenarios. 
The proposed framework shows much better performance than the previous study (DeiT), which validates its generality across various scenarios.
Especially, our approach shows an enlarged performance gap of 92.7\% when the ViT model is randomly initialized on CUB-200, which shows its robustness in the absence of a well-trained model. The `Transfer' results for Flowers-102 are omitted because the dataset becomes a trivial task with the ImageNet pretrained model.

\section{Weakly Supervised Object Localization}

\begin{table}[hbt!]
\centering
\begin{tabular}{|c|cc|}
\hline
IoU threshold       & DeiT & AAL  \\ \hline
0.3                 & 50.3 & 64.5 \\
0.5                 & 20.0 & 30.7 \\
{[}0.3, 0.5{]}      & 35.2 & 48.3 \\
{[}0.3, 0.5, 0.7{]} & 25.0 & 34.2 \\ \hline
\end{tabular}
\caption{Localization accuracy in WSOL task on CUB-200}
\label{tab: wsol}
\end{table}

We additionally evaluate our framework through Weakly Supervised Object Localization (WSOL),
which is frequently used to show the space awareness.
WSOL trains the network model to classify the input image and evaluates the localization of target objects. 
We determine the position of the target objects by averaging the entire attention maps of Eq.~3. 
We measure the localization performance by using the Intersection of Union (IoU) in CUB-200. 

The results are given in Table~\ref{tab: wsol}. 
We refer Choe et al.\footnote{Choe, Oh, Lee, Chun, Akata, and Shim, ``Evaluating Weakly Supervised Object Localization Methods Right'', In CVPR, 2020.} for the evaluation method of WSOL.
We use MaxBoxAcc which measures how the box generated from the attention map overlaps with the ground truth box with a IoU threshold. 
While the default setting is 0.5, we demonstrate results with different IoU thresholds.
The result of multiple IoU threshold indicates average of scores from each threshold in a list.
Compared to DeiT, although the proposed algorithm shows only 1\% top-1 accuracy improvement on CUB-200 in Table \ref{tab:CUBflowers} when the initial models are pretrained on ImageNet,
its localization accuracy is 53.5\% higher than that of DeiT at IoU threshold as 0.5, which validates the space awareness of our knowledge distillation.

\section{Qualitative Results for Attention Maps}

\begin{figure}[ht!]
\centering
\includegraphics[width=0.5\linewidth]{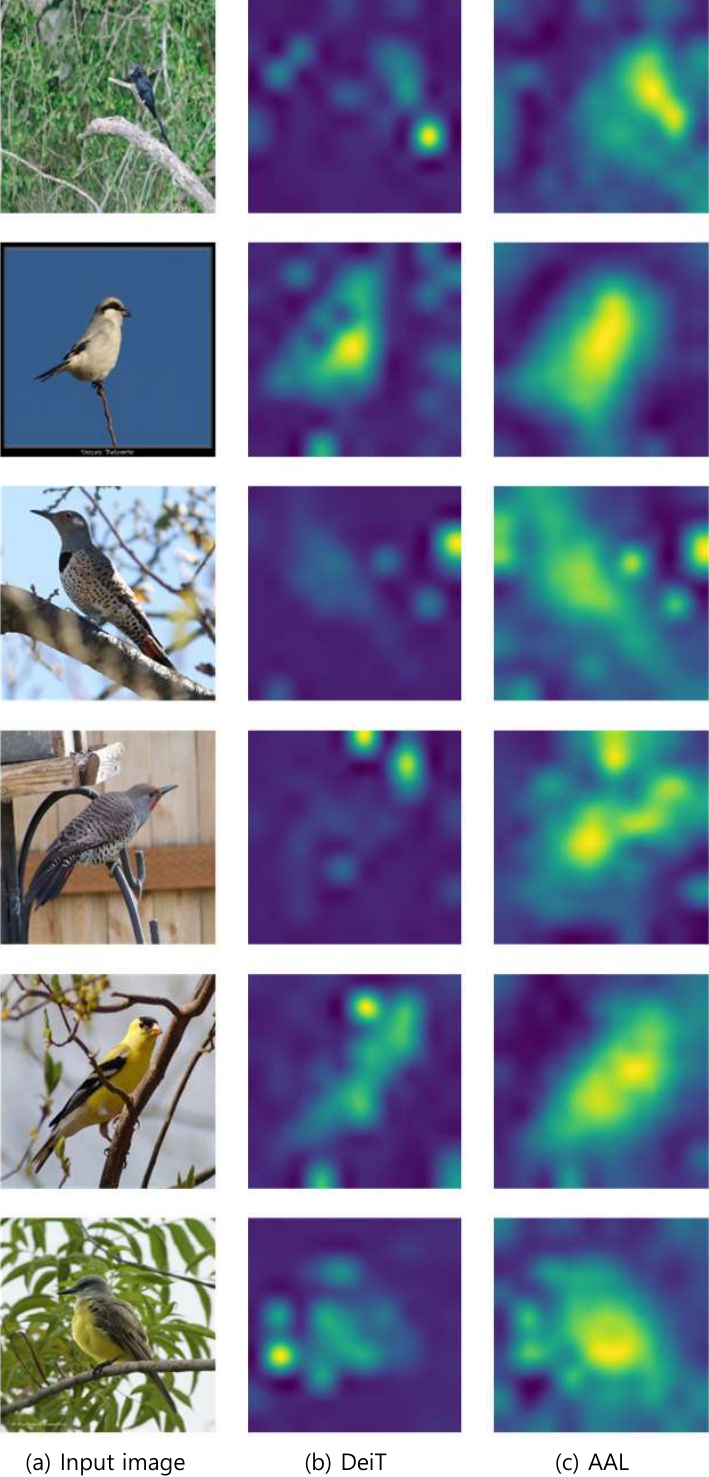}
\caption{Attention maps obtained by a DeiT and AAL after 300 epochs of training on Caltech-UCSD Birds-200-2011. For each row, we illustrate the channel-wise mean of all attention maps extracted by Eq.~3 with the method DeiT and AAL as column (b) and (c), respectively.}
\label{Figure:cub_attention_map}
\end{figure}

For qualitative analysis on the effect of our regularization method to attention maps,
we compare the attention on objects which are acquired by averaging the entire attention maps of Eq.~3. 
For each models, we used same DeiT-S model as baseline but only differed in training strategy, where (b) is DeiT-S with hard distillation and (c) is DeiT-S with AAL.
For training, we used CUB-200-2011 dataset, which only labeled for classification with small number of samples.
After training, we obtained class attention map by the aforementioned processes.
We could tell that the model trained with AAL localizes the objects to be classified better than DeiT, with higher intensity value on the area.
From the result, we could also infer that the inductive bias of CNN-locality is transferred more successfully by observing its attention on object when trained with AAL.
Quantitative results on the attention maps are given in Table~\ref{tab: wsol}.

\section{Epoch-wise Change of Attention Links}

\begin{figure}[hbt!]
\centering
        \includegraphics[width=\linewidth]{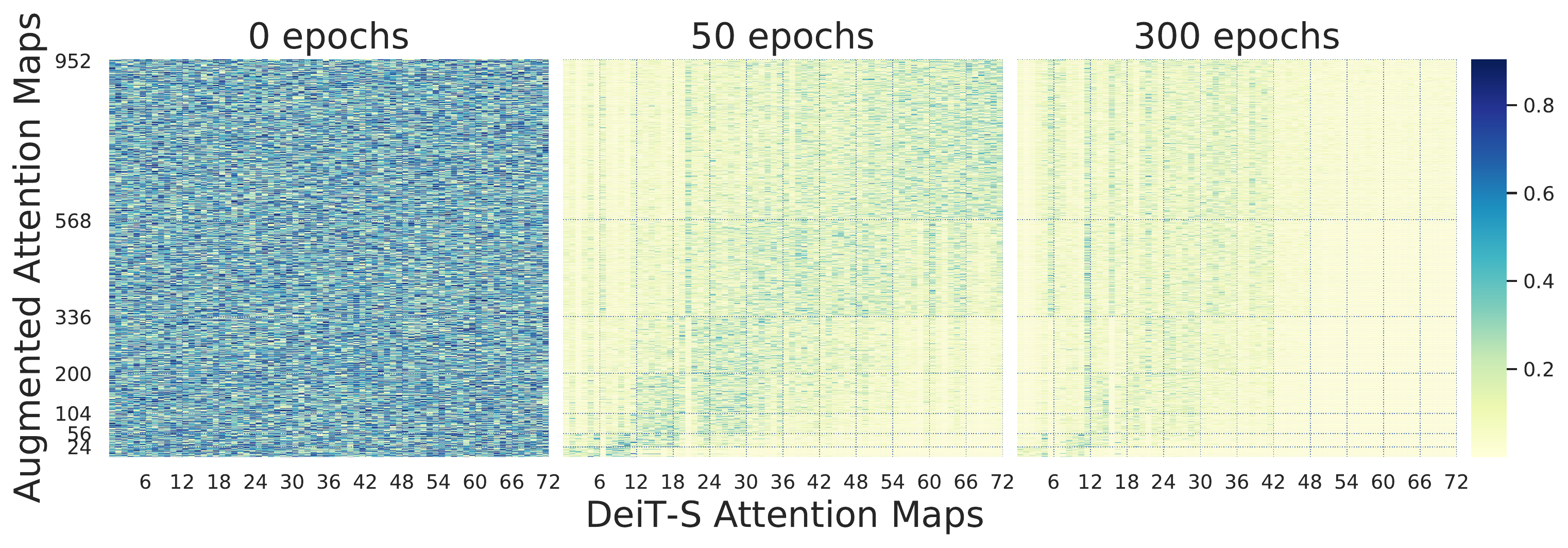}
\caption{Variation of attention link during ImageNet training.}
    \label{Figure:attention link}
\end{figure}

Fig.~\ref{Figure:attention link} shows the transformation of adaptive links during the training on ImageNet. At the end of the training, we observed that high-level heads are disconnected from augmented attention
maps, meaning those are no more regularized by the activation maps.
This indicates that high-level attention heads escape locality and achieve long-range dependency which cannot be acquired by CNNs. From the analysis, we configured β-link
setting to prevent high-level heads from over-regularization.
Table~\ref{table:link} also proves the effectiveness of selective links.

\clearpage
\section{Algorithm for Selective Link Extraction}

\begin{algorithm}[hbt!]
\caption{Selective Link Extraction}\label{selective_link_algorithm}
\hspace*{\algorithmicindent} \textbf{Input}: Trained adaptive link $w_c^{(m, n)}\in W$ where $c\in \{1, ..., C\}$, $m \in \{1,...,M\}$, $n \in \{1, ..., N\}$\\
\hspace*{\algorithmicindent} \textbf{Output}: Selective link $\Omega$
\begin{algorithmic}[0]
\Procedure{Link pruning}{$W$}
 \State $\Omega \gets \{\}$ 
\State $w_{min},\;w_{max} = Min(w_c^{(m,n)}),\;Max(w_c^{(m,n)}) \; \forall \;c,\;m,\;n$
 \For {\texttt{c = 1, ..., C}}
  \For {\texttt{n = 1, ...,  N}}  
   \For {\texttt{m = 1, ..., M}}  
   
    \State $w \gets w_c^{(m, n)}$
    \State $w \gets \frac{w - w_{min}}{w_{max} - w_{min}}$
        \EndFor
   \EndFor
   \EndFor
 \For {\texttt{c = 1, ..., C}}
  \For {\texttt{n = 1, ...,  N}}  

    \If{ $avg(|w_c^{(:,\; n)})|) > \theta_l$ }
     
       \State $\Omega \gets w_c^{(:,\; n)}$
       
   \EndIf
   \EndFor
   \EndFor   

\State \textbf{return} $\Omega$
\EndProcedure
\end{algorithmic}
\label{algorithm}
\end{algorithm}

After training the fully-connected link that connects the original ViT attention map and augmented attention map,
we first normalize the connectivity of every link. Then, if the averaged value of weight that connects each CNN/ViT layer ($C,\,n$, respectively) is larger than a user-defined threshold, we add the links between the layers to a set of selective links. The remaining links are pruned.
While $\theta_l$ is a hyperparameter, we set $\theta_l$ to 0.05 for both the $\alpha$-link and $\beta$-link in our experiments.

\section{Model Flops and Parameter Size}

\begin{table}[h!]
\centering
\begin{tabular}{cccc}
        & DeiT-S       & ConViT-S & AAL             \\ \hline
Params  & 22.4M        & 27.8M    & 22.5M            \\
FLOPs   & 4.27G        & 5.35G    & 4.27G            \\
Runtime &  0.40        & 1.23     & 0.36             \\ \hline
Teacher & RegNetY\_160 & -        & EfficientNet\_B3 \\
Params  & 83.6 M       & -        & 12.2 M           \\
FLOPs   & 15.9G        & -        & 0.98 G           \\ \hline
\end{tabular}
\caption{For estimating the running time, we timed one model process of a batch size of 128 at the training phase in seconds. The time includes the forward pass of teacher CNN for both DeiT-S and AAL. DeiT-S and AAL share similar values in the number of parameters and FLOPs due to the relatively small size of the attention augmentation module.}
\label{tab:comparison}
\end{table}
As shown in Table~\ref{tab:comparison}, our framework shows the fastest runtime at the training phase which includes the inference of teacher CNN among the three comparisons including DeiT and ConViT. This additionally validates the training efficiency of our framework. Furthermore, our approach needs 20.2\% fewer inference FLOPS than ConViT even though they have similar performance.
\clearpage

\section{Robustness to Random Initialization} 

\begin{table}[hbt!]
    \begin{center}
        \centering
        \begin{tabular}{c|ccc}
        \hline
              & Trial I & Trial II & Trial III \\ \hline\hline
        Top-1 & 47.3\%    & 46.5\%     & 47.2\%      \\ 
        Top-5 & 71.9\%    & 71.3\%     & 72.0\%      \\ \hline
        \end{tabular}
        \caption{\textbf{Repeated trials on ImageNet 5\%} We perform several trials with different random seeds. Our algorithm shows consistency even with the various initial parameters.
Due to our limited computation, we run 240 epochs of training in contrast to 300 epochs of training in the default setting.}
        \label{Table 4:random seed}
    \end{center}
\end{table}

\section{Choosing Hyperparameter $\lambda$ and its Decay Rate}

At every experiment, we set the initial value of $\lambda$ to 2000. To justify the value choice, we perform the ablation studies by changing the value of $\lambda$ by 1500, 2000, and 2500. In ImageNet 10\%, we obtain top-1 accuracy of 56.5, 64.7, and 64.6 respectively for $\lambda$ as 1500, 2000, and 2500. We acquire the best performance when $\lambda$ is set to 2000. Since $\lambda$ controls the scale of the regularization loss term, its value highly correlates with the overfitting and the underfitting of the target model. As a result, the value of $\lambda$ should be set to avoid both the overfitting and the underfitting issues, and we empirically found that the value of $\lambda$ as 2000 is a proper choice.

As shown in Fig.~\ref{Figure:attention link}, the ViT attention maps seem similar to the CNN activation maps at the related depth levels, while the relation becomes weakened as training goes on. Accordingly, we designed our approach to reduce the power of the attention-based knowledge distillation loss terms by employing the decay constant of $\lambda$. In addition, we perform the ablation study where the variant is built by fixing the decay constant with $0.99$. In ImageNet 50\%, the variant with the fixed decay constant shows the performance of 77.8\% and 93.8\% for top-1 and top-5 accuracy, respectively, which are less than 78.9\% and 94.5\% of our final model. This result shows that we can tune the decay constant to improve the efficiency of our approach, which would be analyzed in our future work.

\end{document}